		\documentclass[preprint,12pt]{elsarticle}
		
		\usepackage{amsmath, amsfonts, amssymb, amsthm}
		\usepackage{geometry}
		\usepackage{booktabs}
		\usepackage{float}
		\usepackage{multirow}
		\usepackage{rotating}
		
		\journal{journal}
		
\begin{document}
		
		\begin{frontmatter}
		
		
		
		\title{A Significantly Better Class of Activation Functions Than ReLU Like Activation Functions}
		
		
		\author [label1]{Mathew Mithra Noel}
		\author [label2]{Yug Oswal}

		\affiliation[label1]{organization={School of Electrical Engineering, Vellore Institute of Technology}, addressline={Email: mathew.m@vit.ac.in}, country={India}}
		             
		\affiliation[label2]{organization={School of Computer Science and Engineering, Vellore Institute of Technology},   addressline={Email: yoswal071@gmail.com}, country={India}}

		\begin{abstract}
		
		This paper introduces a significantly better class of activation functions than the almost universally used ReLU like and Sigmoidal class of activation functions. Two new activation functions referred to as the Cone and Parabolic-Cone that differ drastically from popular activation functions and significantly outperform these on the CIFAR-10 and Imagenette benchmmarks are proposed. The cone activation functions are positive only on a finite interval and are strictly negative except at the end-points of the interval, where they become zero. Thus the set of inputs that produce a positive output for a neuron with cone activation functions is a hyperstrip and not a half-space as is the usual case. Since a hyper strip is the region between two parallel hyper-planes, it allows neurons to more finely divide the input feature space into positive and negative classes than with infinitely wide half-spaces. In particular the XOR function can be learn by a single neuron with cone-like activation functions. Both the cone and parabolic-cone activation functions are shown to achieve higher accuracies with significantly fewer neurons on benchmarks. The results presented in this paper indicate that many nonlinear real-world datasets may be separated with fewer hyperstrips than half-spaces. The Cone and Parabolic-Cone activation functions have larger derivatives than ReLU and are shown to significantly speedup training.
		
		\end{abstract}

		
\end{frontmatter}
		
		
\section{Introduction}
Since the discovery of ReLU like activation functions \cite{glorot2011deep}, the question of the existence of an even better class of activation functions that differ significantly from both sigmoidal and ReLU like activation functions has remained unanswered \cite{kunc2024three}. This paper answers the above fundamental question in the affirmative by proposing a new class of activation functions. Despite the complexity of deep Artificial Neural Networks (ANNs), each individual neuron in an ANN essentially makes a linear decision by separating its inputs with a single hyperplane. In particular the set of inputs that elicit a positive output from a single neuron is a halfspace.

\subsection{Nature of neuronal decision boundaries}
The output (activation) of a single neuron is given by $a = g(\textbf{w}^T\textbf{x}+b)$, where $g$ is the activation function. The hyperplane boundary associated with a neuron is the set of points: $$H = \{\textbf{x} \in \mathbb{R}^n: \textbf{w}^T\textbf{x}+b =0 \}$$

The set of points for which a neuron produces positive and negative outputs are half-spaces for most populat activation functions. The positive and negative half spaces are defined to be:
$$H_- = \{\textbf{x} \in \mathbb{R}^n: \textbf{w}^T\textbf{x}+b < 0 \}$$
$$H_+ = \{\textbf{x} \in \mathbb{R}^n: \textbf{w}^T\textbf{x}+b > 0 \}$$

Any hyperplane divides it's input space $\mathbb{R}^n$ into 3 connected regions: the positive half-space $H_+$, the negative half-space $H_-$ and an affine-space $H$. The weight vector \textbf{w} points into the positive half-space $H_+$. Fig. X shows illustrates the separation of the input space by a single neuron.  

The cone and the parabolic-cone activation function is defined to be $g(z) = 1-|z-1|$ and $g(z) = z(2-z)$ respectively. A wider class of cone-like activation functions can also be defined: $g(z) = \beta-|z-\gamma|^\alpha$, where $\alpha, \beta$ and $\gamma$ are learnable parameters that affects the shape of the activation function. In contrast to Fig. X, Fig. Y shows the separation of the input space by single neuron with cone-like activation function.

The set of inputs to a neuron that produce a strictly positive output is denoted by $C_+$ and the set of inputs that produce a strictly negative output is denoted by $C_-$.

The set of inputs that exactly produce an output of zero constitutes the decision boundary of the neuron. In particular the decision boundary of a single neuron that produces an output $ a = g(z) = g(\textbf{w}^T\textbf{x}+b)$ is the set $B(g) = \{\textbf{x} \in \mathbb{R}^n: g(\textbf{w}^T\textbf{x}+b) = 0 \}$. Based on the above, the decision boundary of popular activation function like Leaky ReLU, Swish and GELU that zero have a single zero at the origin is a hyperplane.

$$ g(z) = 0 \Longleftrightarrow z = 0 $$   $$ z = \textbf{w}^T\textbf{x}+b = 0 $$ In other words the decision boundary is a single hyperplane ($ B = H =  \{\textbf{x} \in R^n :  \textbf{w}^T\textbf{x}+b = 0 \}$).

On the other hand, the decision boundary of Cone-like activation functions that are zero at the endpoints of a closed interval $[0, \delta]$ consists of two hyperplanes.

$ g(z) = 0 \Longleftrightarrow $  $z = 0 $ OR $z = \delta $  \\\\
$\Longrightarrow $ $\textbf{w}^T\textbf{x}+b = 0  $ OR $ \textbf{w}^T\textbf{x}+b = \delta $ \\

The set of inputs to a Cone-like neuron that produce a strictly positive output is $C_+ = \{\textbf{x} \in \mathbb{R}^n: 0 < \textbf{w}^T\textbf{x}+b < \delta \} $.  Thus $C_+$ for Cone-like neurons is a hyper-strip and not a half-space as is usual for popular activation functions \ref{Table: Definitions}.  

\vspace{2cm} 

\begin{table}[H]
	\centering
	\tiny
	\caption{List of activation functions}
	\label{Table: Definitions}
	\begin{tabular}{|c|c|c|}
		\hline 
		\multirow{2}{*}{\begin{tabular}{c}
				\textbf{Activation}\\\textbf{Function} \end{tabular}} & \textbf{Equation} & \textbf{Range}                            \\ 
		&&\\ \hline
		
		Cone   & $ 1-|z-1| $   & [-$\infty$,1]                                                  \\ \hline
		
		Parabolic-Cone   & $ z(2-z) $   & [-$\infty$,1]                                                  \\ \hline
		
		Parameterized-Cone   & $ 1-|z-1|^\beta $   & [-$\infty$,1]                                                  \\ \hline
		Sigmoid / Softmax   & $\frac{1}{1+e^{-z}}$   & [0,1]                                                          \\ \hline
		
		Tanh & $\tanh{(z)}$      & $\left[-1,1\right]$                        \\ \hline
		
		LiSHT               &      $z\tanh{(z)}$    &  ($-\infty$,$\infty$)          \\ \hline
		Softplus            &   $\ln{(1+e^z)}$          &  [0,$\infty$)   \\ \hline
		ReLU &    $\max{(0,z)}$     & [0,$\infty$)    \\ \hline
		Leaky ReLU          &  $\left\{\begin{array}{ll}
			0.01z & z<0;  \\
			z & z\geq 0; \\ 
		\end{array} \right.$  &  ($-\infty$, $\infty$)         \\ \hline
		GELU&	$ 0.5z(1+\tanh{(\sqrt{\frac{2}{\pi}}z
			+0.044715z^3)})$
		&[-0.5,$\infty$)          \\ \hline
		SELU &   $\left\{\begin{array}{ll}
			\lambda z &  z\geq 0;  \\
			\lambda\alpha (e^z-1) &  z< 0. \\
			\alpha\approx 1.6733 & \lambda \approx 1.0507 
		\end{array} \right.$       &  [$-\lambda\alpha$,$\infty$)                             \\ \hline
		Mish                &     $z\tanh{(\ln{(1+e^z)})}$     & [-0.31,$\infty$)        \\ \hline
		Swish               &  $\frac{z}{1+e^{-z}}$      &  [-0.5,$\infty$)           \\ \hline			
		ELU & $\left\{\begin{array}{ll}
			z &  z\geq 0;  \\
			(e^z-1) &  z< 0. \\ 
		\end{array} \right.$ &[-1,$\infty$) \\ \hline

	\end{tabular}
\end{table}	

\begin{figure}[H]  
	\centering
	\includegraphics[width=12cm]{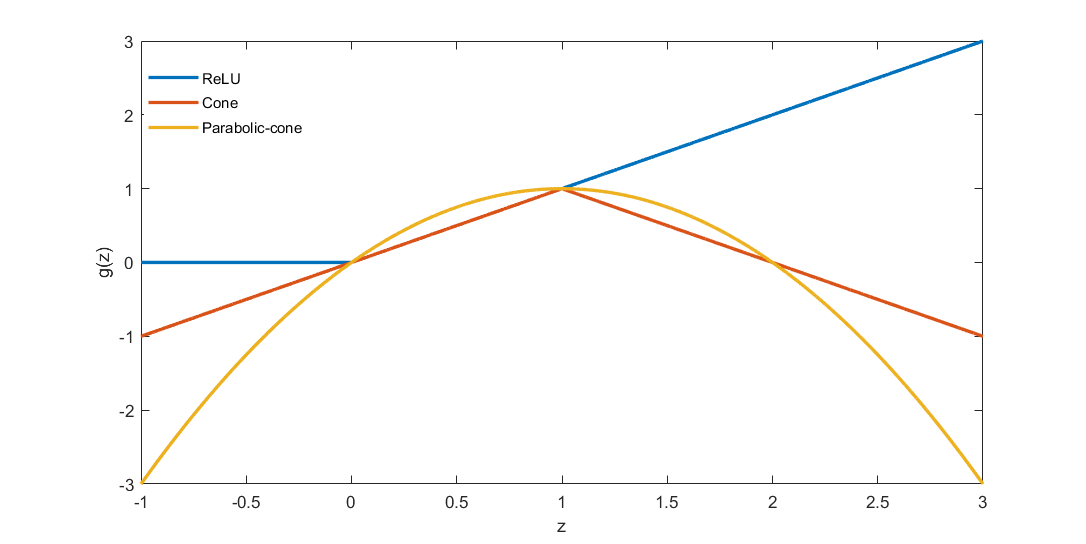}
	\caption{Comparison of ReLU with Cone and Parabolic-Cone activation functions. The set of inputs that provide a strictly positive output for Cone and Parabolic-Cone activation functions is a finite interval $(0,2)$ as apposed to $(0,\infty)$ for ReLU.}
	\label{PlotActivations}
\end{figure}

\begin{figure}[H]  
	\centering
	\includegraphics[width=12cm]{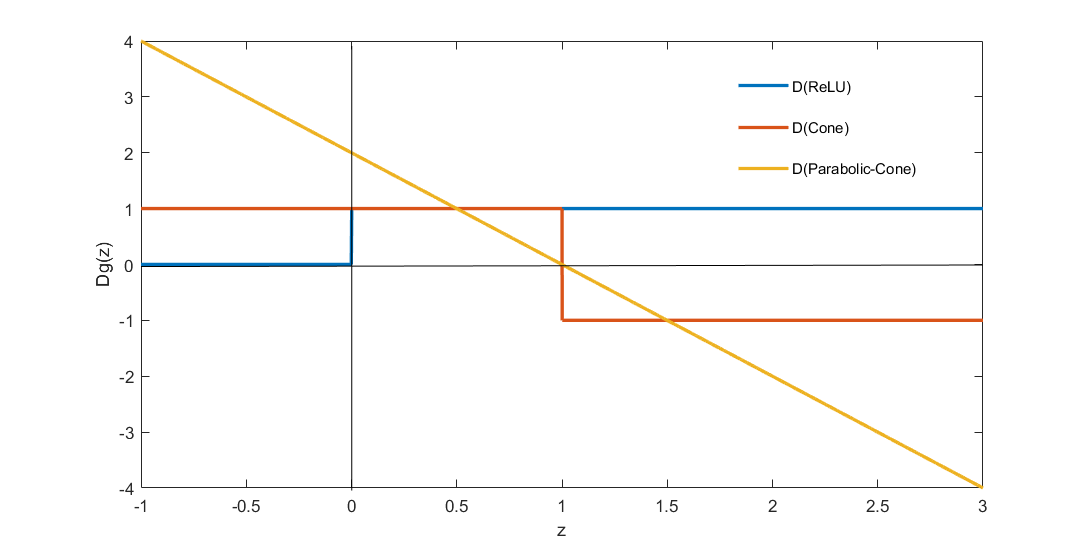}
	\caption{A Comparison of the first derivatives of different activation functions. Cone-like activation functions never saturate and have larger derivative values for most inputs.}
	\label{PlotDerivatives}
\end{figure}
 
\begin{figure}[H]  
	\centering
	\includegraphics[width=12cm]{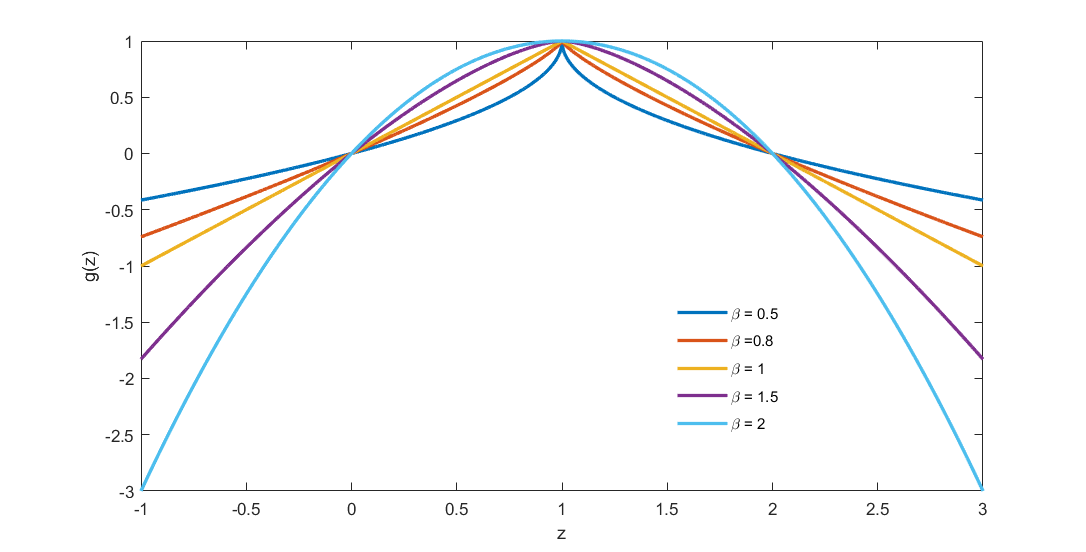}
	\caption{Variation in the shape of the Parameterized-Cone activation with parameter $\beta$.}
	\label{Parametrized-Cone}
\end{figure}

Fig. \ref{PlotActivations} compares the Cone and Parabolic-Cone activation functions with ReLU. Fig. compares the derivatives of different activation functions. Cone-like activation functions never have small or zero derivative for any input. Cone-like activation functions also have larger derivative values than ReLU for most inputs facilitating faster learning. A parameterized version of the Cone activation is shown in Fig. \ref{Parametrized-Cone}.

\subsection{Halfspaces versus Hyper-strips}
Since hyper-strips are narrower compared to infinitely wide halfspaces, fewer hyper-strips are needed to accurately partition the inputs space into different classes. Fig. \ref{Annulus} below illustrated how a simple two layer ANN with just 2 hidden layer Cone neurons and a  single sigmoidal neuron can learn a complex linearly non-separable dataset.

\begin{figure}[H]  
	\centering
	\includegraphics[width=8cm]{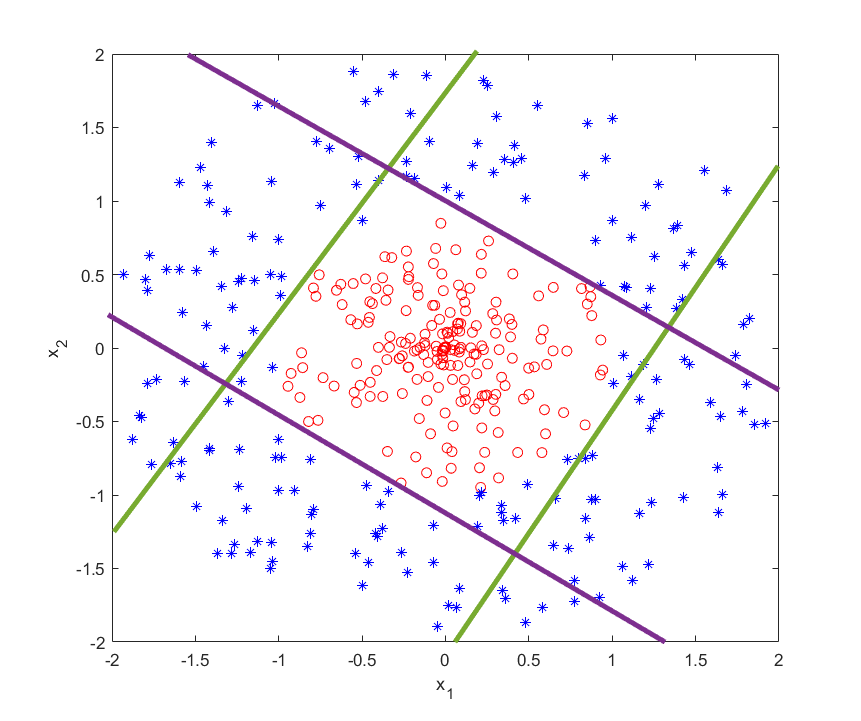}
	\caption{Only two hyper-strips are needed to accurately partition this dataset. Two neurons with Cone or Parabolic-Cone can be used to learn the 2 hyper-strips. However 4 ReLU or sigmoidal neurons will be needed to learn 4 hyperplane boundaries.}
	\label{Annulus}
\end{figure}

\begin{figure}[H]  
	\centering
	\includegraphics[width=8cm]{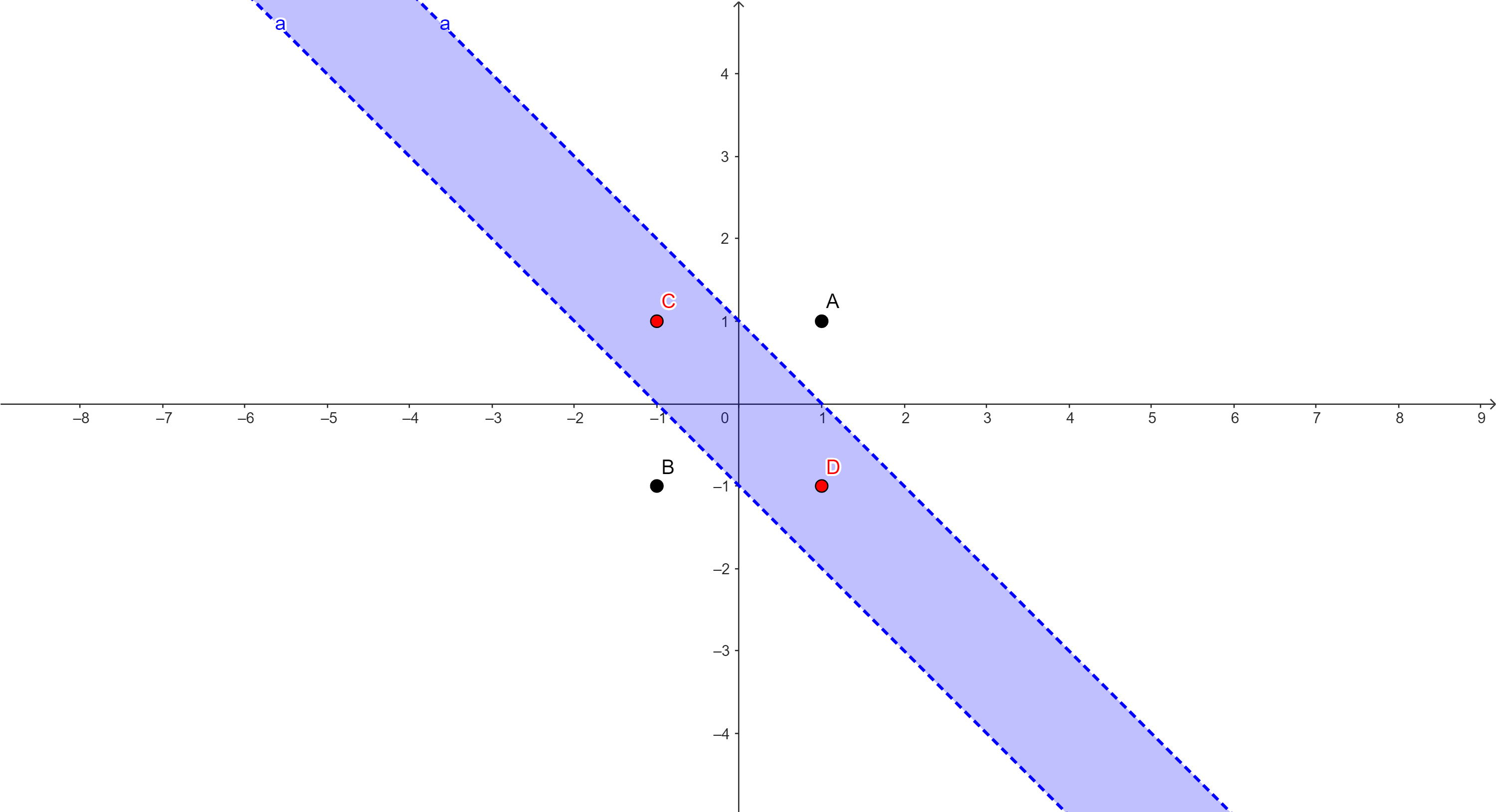}
	\caption{The classic XOR problem can be solved with a single neuron with Cone activation, since $C_+$ is a hyper-strip for Cone-like neurons.}
	\label{XOR}
\end{figure}

Fig. \ref{XOR} illustrates how the classic XOR problem can be solved with a single neuron with Cone-like activation function. The main contributions of this paper are:
\begin{itemize}

		    \item A new class of activation functions significantly better than ReLU like activation functions is proposed
		    
		    \item The proposed activation functions learn a hyper-strip instead of a halfspace to separate the class of positive inputs.
		    
		    \item Using hyper-strips instead of halfspaces to separate the class of positive inputs is shown to result in smaller ANNs
		  		   
\end{itemize}

\section{Results: Performance comparison on benchmark datasets}
\label{section:results}

In the following, Cone-like activation functions are compared  with most popular activation functions on the CIFAR-10 \cite{alex2009learning} and Imagenette \cite{Imagenette} benchmarks. Appendix-I and Appendix-II show the CNN architectures used with different benchmarks. A standard CNN architecture consisting of convolutional layers followed by fully connected dense layers was used. The features computed by the convolutional filters are input to a single dense layer with the activation function being tested. The output in all cases consisted of a Softmax layer. Tables below show the average results obtained over 5 independent trials to account for variation in performance due to random initialization. Adam optimizer with a learning rate of 10-4 and standard momentum with both $\beta_1$ and $\beta_2$ was used to train the models. All models were trained for 30 epochs with Categorical Cross-Entropy loss. Tables \ref{CIFAR10_1}, \ref{CIFAR10_2} and \ref{CIFAR10_3} clearly show that the Cone and Parabolic-Cone activation functions achieve higher accuracies on the CIFAR-10 benchmark with significantly fewer neurons.  

 \begin{table}[H]
	\centering
	\scriptsize
	\caption{Performance comparison of different activation functions on the CIFAR-10 dataset \\ with a single fully-connected layer composed of 64 neurons.}
	\label{CIFAR10_1}
	\begin{tabular}{|c|c|c|c|c|c|}
		\hline
		\begin{tabular}{c}
			\textbf{Activation} \\ \textbf{Function} \end{tabular} & 
		\begin{tabular}{c} \textbf{Mean Test}\\ 
			\textbf{Accuracy} \end{tabular} & \begin{tabular}{c} \textbf{Median Test}\\ \textbf{Accuracy}\end{tabular} & \begin{tabular}{c} \textbf{Std. Dev. Test}\\ \textbf{Accuracy}\end{tabular} & \begin{tabular}{c} \textbf{Best Test}\\ \textbf{Accuracy}\end{tabular}  & \begin{tabular}{c} \textbf{Worst Test}\\ \textbf{Accuracy} \end{tabular}\\ \hline
		
		ReLU      & 0.7196 & 0.7208 & 0.0044 & 0.7234 & 0.7120 \\ \hline
		Leaky ReLU  &   0.7247 & 0.7234 & 0.0035 & 0.7303 & 0.7209  \\ \hline
		Cone  & 0.742728 & 0.7427 & 0.0046 & 0.7495 & 0.7377  \\ \hline
	    Parabolic-Cone  &  0.7510  & 0.7499 & 0.006 & 0.7586 & 0.7446  \\ \hline
				
	\end{tabular}
\end{table}

 \begin{table}[H]
	\centering
	\scriptsize
	\caption{Performance comparison of different activation functions on the CIFAR-10 dataset \\ with a single fully-connected layer composed of 32 neurons.}
	\label{CIFAR10_2}
	\begin{tabular}{|c|c|c|c|c|c|}
		\hline
		\begin{tabular}{c}
			\textbf{Activation} \\ \textbf{Function} \end{tabular} & 
		\begin{tabular}{c} \textbf{Mean Test}\\ 
			\textbf{Accuracy} \end{tabular} & \begin{tabular}{c} \textbf{Median Test}\\ \textbf{Accuracy}\end{tabular} & \begin{tabular}{c} \textbf{Std. Dev. Test}\\ \textbf{Accuracy}\end{tabular} & \begin{tabular}{c} \textbf{Best Test}\\ \textbf{Accuracy}\end{tabular}  & \begin{tabular}{c} \textbf{Worst Test}\\ \textbf{Accuracy} \end{tabular}\\ \hline
		
		ReLU      & 0.7196 & 0.6893 & 0.0061 & 0.6952 & 0.6807 \\ \hline
		Leaky ReLU  &   0.7052 & 0.7040 & 0.0113 & 0.7238 & 0.6953  \\ \hline
		Cone  & 0.7291 & 0.7260 & 0.0062 & 0.7388 & 0.7236  \\ \hline
		Parabolic-Cone  &  0.7439  & 0.7428 & 0.0065 & 0.7538 & 0.7378  \\ \hline
		
	\end{tabular}
\end{table}

 \begin{table}[H]
	\centering
	\scriptsize
	\caption{Performance comparison of different activation functions on the CIFAR-10 dataset \\ with a single fully-connected layer composed of 10 neurons.}
	\label{CIFAR10_3}
	\begin{tabular}{|c|c|c|c|c|c|}
		\hline
		\begin{tabular}{c}
			\textbf{Activation} \\ \textbf{Function} \end{tabular} & 
		\begin{tabular}{c} \textbf{Mean Test}\\ 
			\textbf{Accuracy} \end{tabular} & \begin{tabular}{c} \textbf{Median Test}\\ \textbf{Accuracy}\end{tabular} & \begin{tabular}{c} \textbf{Std. Dev. Test}\\ \textbf{Accuracy}\end{tabular} & \begin{tabular}{c} \textbf{Best Test}\\ \textbf{Accuracy}\end{tabular}  & \begin{tabular}{c} \textbf{Worst Test}\\ \textbf{Accuracy} \end{tabular}\\ \hline
		
		ReLU      & 0.6157 & 0.6110 & 0.0140 & 0.6317 & 0.5993 \\ \hline
		Leaky ReLU  &   0.6292 & 0.6329 & 0.0110 & 0.6405 & 0.6174  \\ \hline
		Cone  & 0.6844 & 0.6783 & 0.0122 & 0.6999 & 0.6714  \\ \hline
		Parabolic-Cone  &  0.6998  & 0.6997 & 0.0064 & 0.7088 & 0.6920  \\ \hline
		
	\end{tabular}
\end{table}

Tables \ref{Imagenette_1}, \ref{Imagenette_2} and \ref{Imagenette_3} show that the Cone and Parabolic-Cone activation functions achieve overall higher accuracies on the Imagenette benchmark when the number of neurons is reduced.

\begin{table}[H]
	\centering
	\scriptsize
	\caption{Performance comparison of different activation functions on the Imagenette benchmark \\ with a single fully-connected layer composed of 64 neurons.}
	\label{Imagenette_1}
	\begin{tabular}{|c|c|c|c|c|c|}
		\hline
		\begin{tabular}{c}
			\textbf{Activation} \\ \textbf{Function} \end{tabular} & 
		\begin{tabular}{c} \textbf{Mean Test}\\ 
			\textbf{Accuracy} \end{tabular} & \begin{tabular}{c} \textbf{Median Test}\\ \textbf{Accuracy}\end{tabular} & \begin{tabular}{c} \textbf{Std. Dev. Test}\\ \textbf{Accuracy}\end{tabular} & \begin{tabular}{c} \textbf{Best Test}\\ \textbf{Accuracy}\end{tabular}  & \begin{tabular}{c} \textbf{Worst Test}\\ \textbf{Accuracy} \end{tabular}\\ \hline
		
		ReLU      & 0.852229 & 0.850701 & 0.004028 & 0.859363 & 0.849682 \\ \hline
		Leaky ReLU  &   0.852688 & 0.851465 & 0.006892 & 0.863185 & 0.844076  \\ \hline
		Cone  & 0.852994 & 0.852739 & 0.007127 & 0.863185 & 0.843057  \\ \hline
		Parabolic-Cone  &  0.847949  & 0.85172 & 0.007872 & 0.855541 & 0.838471  \\ \hline
		
	\end{tabular}
\end{table}

\begin{table}[H]
	\centering
	\scriptsize
	\caption{Performance comparison of different activation functions on the Imagenette benchmark \\ with a single fully-connected layer composed of 32 neurons.}
	\label{Imagenette_2}
	\begin{tabular}{|c|c|c|c|c|c|}
		\hline
		\begin{tabular}{c}
			\textbf{Activation} \\ \textbf{Function} \end{tabular} & 
		\begin{tabular}{c} \textbf{Mean Test}\\ 
			\textbf{Accuracy} \end{tabular} & \begin{tabular}{c} \textbf{Median Test}\\ \textbf{Accuracy}\end{tabular} & \begin{tabular}{c} \textbf{Std. Dev. Test}\\ \textbf{Accuracy}\end{tabular} & \begin{tabular}{c} \textbf{Best Test}\\ \textbf{Accuracy}\end{tabular}  & \begin{tabular}{c} \textbf{Worst Test}\\ \textbf{Accuracy} \end{tabular}\\ \hline
		
		ReLU      & 0.844178 & 0.841783 & 0.00504 & 0.851465 & 0.838981 \\ \hline
		Leaky ReLU  &   0.851006 & 0.846879 & 0.005905 & 0.859108 & 0.846624  \\ \hline
		Cone  & 0.85228 & 0.851465 & 0.003751 & 0.857834 & 0.848408  \\ \hline
		Parabolic-Cone  &  0.849121  & 0.849682 & 0.004422 & 0.855796 & 0.844076  \\ \hline
		
	\end{tabular}
\end{table}

\begin{table}[H]
	\centering
	\scriptsize
	\caption{Performance comparison of different activation functions on the Imagenette benchmark\\ with a single fully-connected layer composed of 10 neurons.}
	\label{Imagenette_3}
	\begin{tabular}{|c|c|c|c|c|c|}
		\hline
		\begin{tabular}{c}
			\textbf{Activation} \\ \textbf{Function} \end{tabular} & 
		\begin{tabular}{c} \textbf{Mean Test}\\ 
			\textbf{Accuracy} \end{tabular} & \begin{tabular}{c} \textbf{Median Test}\\ \textbf{Accuracy}\end{tabular} & \begin{tabular}{c} \textbf{Std. Dev. Test}\\ \textbf{Accuracy}\end{tabular} & \begin{tabular}{c} \textbf{Best Test}\\ \textbf{Accuracy}\end{tabular}  & \begin{tabular}{c} \textbf{Worst Test}\\ \textbf{Accuracy} \end{tabular}\\ \hline
		
		ReLU      & 0.774573 & 0.773758 & 0.006244 & 0.78344  & 0.767898 \\ \hline
		Leaky ReLU  &   0.822981 & 0.824713 & 0.006718 & 0.82879 & 0.811465  \\ \hline
		Cone  & 0.830981 & 0.829809 & 0.002532 & 0.834904 & 0.82879  \\ \hline
		Parabolic-Cone  &  0.842089  & 0.839745 & 0.007008 & 0.852229 & 0.835159  \\ \hline
		
	\end{tabular}
\end{table}

\begin{figure}[H]  
	\centering
	\includegraphics[scale=0.5]{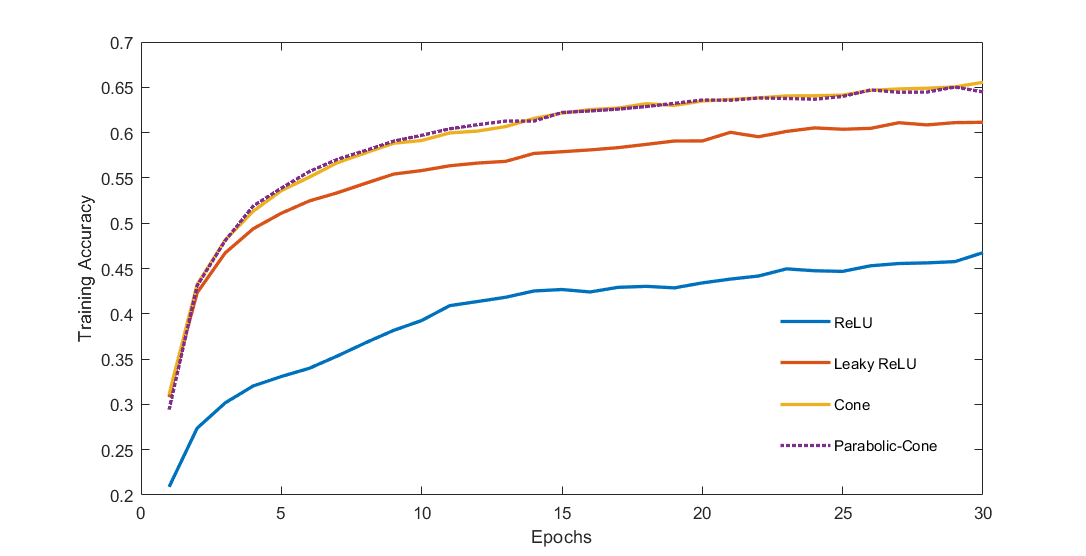}
	\caption{Training curves with different activation functions on CIFAR-10 with a single layer of 10 dense neurons.}
	\label{CIFAR10_Training}
\end{figure}

The Cone and the Parabolic-Cone activation functions proposed in this paper also significantly speedup training (Fig. \ref{CIFAR10_Training}). The faster training can be attributed to the larger derivative values for Cone-like activation functions for most inputs compared to other activation functions. The fundamental question of whether activation functions with even larger derivatives will train faster or lead to the exploding gradient problem remains unanswered.

\section{Conclusion}
ReLU like activation function differ drastically from sigmoids and significantly outperform sigmoidal activation functions and allows deep ANNs to be trained by alleviating the "Vanishing Gradient Problem." Thus a fundamental question in the field of neural networks is the question of whether an even better class of activation functions that differ substantially from ReLU like and sigmoidal activation functions exist. Inspired by the fact that hyper-strips allow smaller cuts to be made in the input space than halfspaces, this paper proposed a new class of cone-like activation functions. Cone-like activation functions use a hyper-strip to separate $C_+$ (the set of inputs that elicit a positive output) from other inputs. Since $C_+$ is a hyper-strip the XOR function can be learned by a single neuron with Cone-like activation functions. The paper showed that many nonlinearly separable datasets can be separated with fewer hyper-strips than half-spaces resulting in smaller ANNs. ANNs with Cone-like activation functions are shown to achieve higher accuracies with significantly fewer neurons on CIFAR-10 and Imagenette benchmarks. Results indicate that Cone-like activation functions with larger derivatives than ReLU-like activation functions speedup training and achieve higher accuracies.

\pagebreak

\appendix{Appendix I: CNN architecture for CIFAR-10}

\begin{figure}[H]
	\centering
	\begin{sideways}
	\includegraphics[scale=0.7]{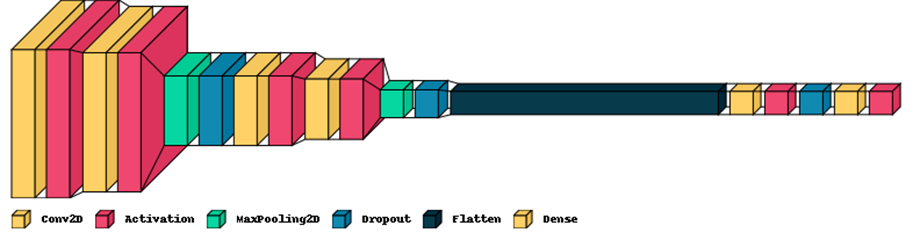}
    \end{sideways}
\end{figure}

\pagebreak
\begin{figure}[H]
	\centering
	\includegraphics[scale=0.8]{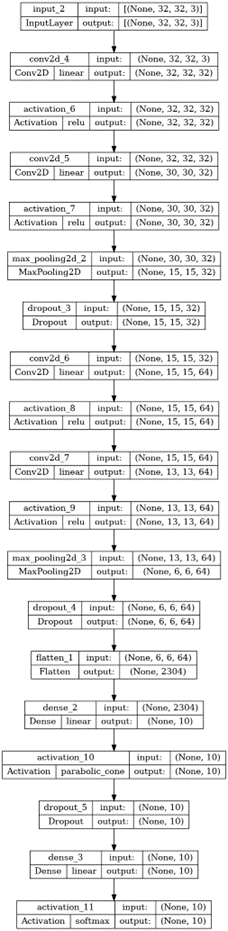}
\end{figure}

\pagebreak

\appendix{Appendix II: CNN architecture for Imagenette}
\begin{figure}[H]
	\centering
	\begin{sideways}
	\includegraphics[scale=0.7]{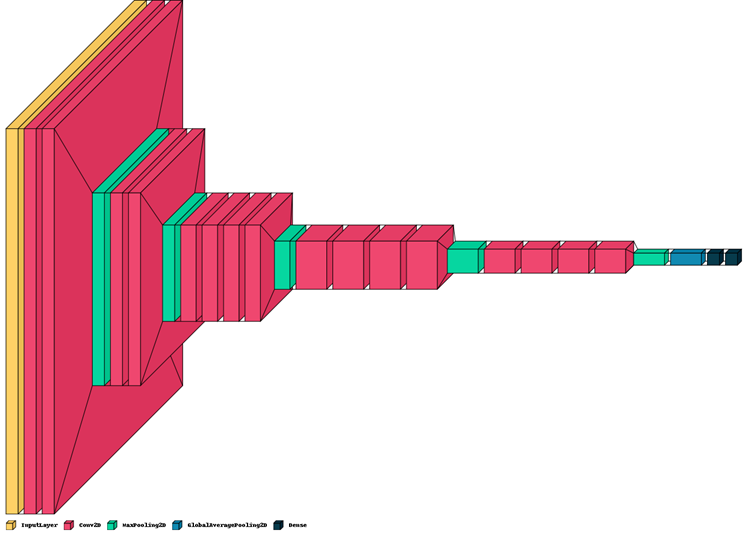}
    \end{sideways}
\end{figure}

\begin{figure}[H]
	\centering
	
	\includegraphics[scale=0.8]{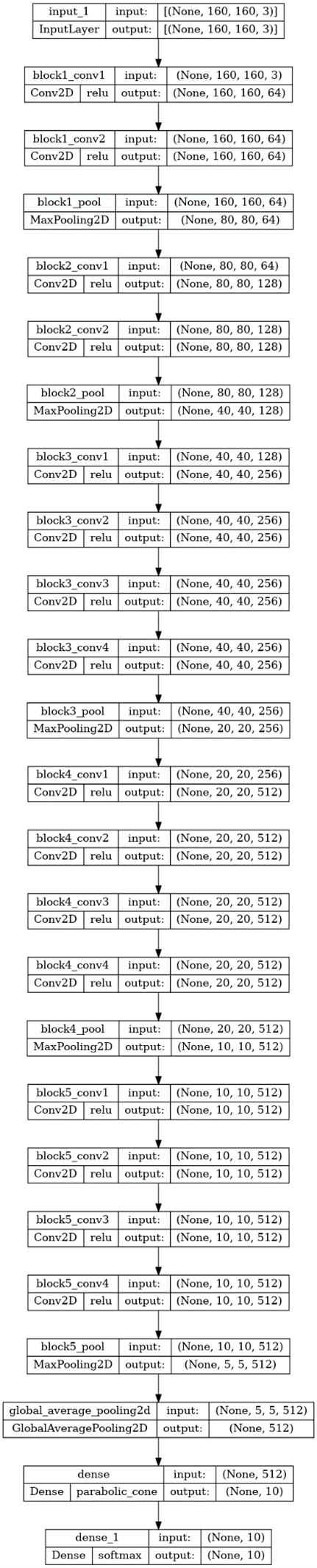}
	
\end{figure}

\bibliographystyle{elsarticle-num}
\bibliography{bibliography.bib}

\end{document}